# Voting-based Multimodal Automatic Deception Detection

*Lana Touma, Mohammad Al Horani, Manar Tailouni, Anas Dahabiah, Khloud Al Jallad
Arab International University
Damascus Syria

*Corresponding Author: 201711039@aiu.edu.sy

**Abstract**

Automatic Deception Detection has been a hot research topic for a long time, using machine learning and deep learning to automatically detect deception, brings new light to this old field. In this paper, we proposed a voting-based method for automatic deception detection from videos using audio, visual and lexical features. Experiments were done on two datasets, the Real-life trial dataset by Michigan University and the Miami University deception detection dataset. Video samples were split into frames of images, audio, and manuscripts. Our Voting-based Multimodal proposed solution consists of three models. The first model is CNN for detecting deception from images, the second model is Support Vector Machine (SVM) on Mel spectrograms for detecting deception from audio and the third model is Word2Vec on Support Vector Machine (SVM) for detecting deception from manuscripts. Our proposed solution outperforms state of the art. Best results achieved on images, audio and text were 97%, 96%, 92% respectively on Real-Life Trial Dataset, and 97%, 82%, 73% on video, audio and text respectively on Miami University Deception Detection.

***Keywords:*** deception detection, trustworthiness, lie detection, Mu3d dataset, real life trial dataset

## 1. Introduction:

In recent years, many research works were done on automated deception detection stating that it may be an efficient solution for different problems such as deception in job interviews and court room trials.

Lying has a huge effect in our day to day lives. For example, in court trials where it could lead to falsely accusing the innocents and freeing the guilty. Also, in job interviews where hiring the wrong employees could prove detrimental to a company's success. This is why it is important to get an accurate decision on whether the person is telling the truth or not in such situations.

Traditional methods for deception detection include analyzing heart beats shifts in posture gaze aversion and limb movements. A study conducted in 2003 [1] shows that liars tell far fewer interesting stories than truth-tellers and that liars also make worse impressions and their demeanor is less calm in general the stories they tell seem more perfect and often contain unrealistic situations.

One of the most popular ways of detecting deception is polygraph or lie-detector machines which monitor heartbeat and physical cues. An article published by the co-inventor of the modern polygraph, L Keeler [2] mentions that the device consists of three units, one recording continuously and quantitatively the subject's blood pressure and pulse, one giving a duplicate blood-pressure pulse curve taken from some other part of the subject's body, and the third recording respiration. However, the device's success in revealing deception and guilt in criminal suspects is largely due to the psychological impact of such tests with an estimated 75% of convicted suspects being tested confessing their crimes. With that being said, this approach is impractical in most cases because it requires the use of skin-contact devices and a human expert's opinion to obtain accurate measurements and interpretations.

Considering the drawbacks of traditional methods of deception detection, automating the process of deception detection has been a hot research topic in recent years.
An article published in 2019 with the title "Can a Robot Catch You Lying? A Machine Learning System to Detect Lies During Interactions" [3] discusses the potential for robots to autonomously detect deception and aid in human interactions. The study involved showing participants videos of robberies and then interrogating them about what they saw, with half of their responses being true and half being false. The study found that there were strong similarities in participants' behavior when interacting with a human and a humanoid robot, and that certain behavioral variables could be used as markers of deception. The results suggest that robots could effectively detect lies in human-robot interactions using these markers. The article does not provide a detailed list of all the markers of deception that were used in the study. However, it mentions that behavioral variables such as eye movements, time to respond, and eloquence were measured during the task and were found to be valid markers of deception in both human-human and human-robot interactions. Other potential markers of deception could include changes in vocal pitch, facial expressions, and body language.

A well-known book by Paul Ekman [4], a pioneer in deception detection research, covers clues of detecting lies based on verbal, vocal and facial cues. The book is titled "Telling Lies: Clues to Deceit in the Marketplace, Politics, and Marriage", and Paul's main takeaways wer :

- Micro expressions - brief, involuntary facial expressions - can reveal when a person is lying or experiencing a negative emotion.
- Baseline statements are useful to compare changes in a person's vocal and facial cues when they are being deceptive.
  Multiple clues from verbal, vocal and facial cues together are more reliable indicators of deception than any single cue alone.

Overall, the use of automated deception detection could provide a more accurate and practical solution for detecting lies in different situations. By extracting various features from data including visual features such as hand movements and facial features or acoustic features such as tone and pitch or lexical features by analyzing the spoken text and then passing those features through different machine learning models, researchers have concluded that it's possible to automatically detect deception from videos and obtain accurate results.

## 2. Related works:

Automatic deception detection is still a new research domain as the first research paper in automatic deception detection from videos using data science was done in 2015. There are two basic types of features that researches extract from videos in this domain, Verbal features (text and audio) and non-verbal features (images). Deep learning and machine learning models were applied on each type of features. Moreover, Studies on multi-model approaches have shown that using features from multiple modalities enhances the detection of deceptive behaviors to a significant degree when compared to using only one modality at a time. [5]

Table 1 is a comparison between state-of-the-art.

| Ref | Year | Dataset(s) | Verbal features | Non-verbal features | Models | Results |
|---|---|---|---|---|---|---|
| **[6]** | 2015 | Real-life Trial Dataset | **Lexical:** Unigrams and bigrams derived from the bag-of-words representation from videos transcripts. | Two broad categories: -Facial features, -Hand gestures. | Decision Tree, Random Forest. | Accuracies in the range of 60-75%. **Highest accuracy is 75.20%** on Decision Tree using all features. |
| **[7]** | 2019 | Real-life Trial Dataset | **Lexical:** Simple weighted unigram features from bag-of-words + emotional information from SenticNet.<br><br>**Acoustic:** basic features like Mel-frequency coefficient, harmonics-to-noise ratio, jitter was extracted using openSMILE. | Facial features: movements of the eyebrows, mouth and eyes, derived from OpenFace library. | Support Vector Machine (SVM) | Accuracy on text modality (66.12%) is higher than previous works. And a lower one in visual modality (67.20%).<br><br>**Highest Accuracy is 78.95%** using feature-level fusion. |
| [8] | 2017 | Real-life Trial Dataset | **Lexical:**<br>CNN model On 300-dimensional GloVe.<br>**Acoustic:**<br>Audio features, such as pitch and voice intensity, are extracted using widely used open-source software openSMILE. | Visual features are extracted from the videos using a 3D CNN. | CNN for each model | Audio based Model:<br>87.5%<br><br>Automated extracted textual cues<br>83.78%<br><br>Visual based 3D deep CNN<br>78.57%<br><br>Early fusion on Text + Audio + Video<br>**96.42%** |
| [9] | 2022 | Real-life Trial Dataset | **Lexical:**<br>Unigrams from Bag-of-words representation + features derived from the Linguistic Inquire Word Count lexicon.<br>**Acoustic:**<br>Pitch, estimated by obtaining the fundamental frequency (f0) of the defendants' speech using the STRAIGHT toolbox.<br>+ Silence and Speech Histograms obtained by running a voice activity detection algorithm. | Facial Action Units (FACS). Using the OpenFace library with the default multi-person detection model to obtain 18 binary indicators of Action Units (AUs). | Support Vector Machine (SVM), Random forest, Feed Forward Neural Network with two hidden layers and SoftMax activation function | **Visual:**<br>Accuracy is 61.5% Using Random Forest<br>**Linguistic**:<br>Accuracy is 71.7%<br>Using a two hidden layers convolutional neural network (100 and 500 nodes for the hidden layers)<br>**Acoustic:**<br>Accuracy is 63.28%<br>Using Random Forest. |

**TABLE 1: RELATED WORKS**

By reviewing previous works from Table 1, we see that Verónica et al. [6] presented a novel dataset consisting of 121 deceptive and truthful clips, from real court trial videos. They used unigrams and bigrams derived from the bag-of-words representation of the video transcripts, and manually annotated the videos for several gestures that were then used to extract non-verbal features such as facial displays and hand gestures. They then built classifiers relying on individual or combined sets of verbal and non-verbal features, achieving accuracies in the range of 60-75%." on real-life trial dataset. This is stated to be the first work to automatically detect deception using both verbal and non-verbal features extracted from real trial recordings.

Jaiswal et al. [7] analyzed both the movement of facial features and the acoustic patterns of the witness and performed a lexical analysis on the spoken words. They improved on the previous study by using a Support Vector Machine (SVM) model and achieved a higher accuracy of 78.95% on the real-life trial dataset.

Gogate et al [8] showed that a deep learning approach improved results. They achieved 96.42% accuracy on real-life trial dataset using early fusion and accuracies of 87.5%, 83.78% ,78.57% on audio, text and video respectively. This was also stated to be the first time use of audio cues for deception detection.

M. Umut Şen et al [9] did the most recent study, they experimented with linguistic features derived from the text transcripts that have been previously found to correlate with deception cues, extracting unigrams from the bag of words representation of each transcript and features derived from the Linguistic Inquire Word Count (LIWC) lexicon. They also extracted a set of visual features consisting of assessments of several facial movements described as Facial Action Units, these features denote the presence of facial muscle movements that are commonly used for describing and classifying expressions. The OpenFace library was used with the default multi-person detection model to obtain 18 binary indicators of Action Units (AUs) for each frame in the videos. Finally, for acoustic features they used pitch, estimated by obtaining the fundamental frequency (f0) of the defendants' speech using the STRAIGHT toolbox, plus silence and Speech Histograms obtained by running a voice activity detection algorithm.

Results showed that the best result is 72.88% on real-life trial dataset, obtained with the score-level combination, and the NN classifier. They also present a human deception detection study where they evaluate the human capability of detecting deception. Results show that the system they built outperforms the average human capability of identifying deceit.

## 3. Datasets:

We conducted our experiments on two datasets, the real-life trial dataset by the university of Michigan [6] and the Miami University Deception Detection Dataset (MU3D). [10] In this section we will explain about both of them.

### *3.1 Real-life trial dataset*

To the best of our knowledge, this dataset is used as a baseline for deception detection in real-life videos which is why we chose it. The dataset consists of 121 videos including 61 deceptive and 60 truthful clips taken from various real-life trial videos where some restrictions were imposed for instance the witness must be clearly identified in the video and their face has to be sufficiently visible for most of the clip. Also, the visual quality has to be clear enough to discern the facial expressions. Lastly, the voice quality should be clear enough to hear the voice and understand what the person is saying. All the video clips were transcribed via crowd sourcing using Amazon Mechanical Turk. The transcribers were asked to insert repetitive words or fillers such as "um", "ah", "uh" and to indicate deliberate silence using ellipsis. Incoming transcriptions were manually checked to avoid spam and ensure quality. The final transcription set consisted of 8,055 words, with an average of 66 words per transcription.

### *3.1 Miami University Deception Detection Dataset (MU3D)*

A dataset resource published by the university of Miami available for free, featuring people telling truthful and deceptive stories. Transcriptions were done by trained researcher assistants and assessed by naïve raters and include all words and sounds indicating silence such as ‘um’, ‘uh’ but they don’t contain things like coughs, laughs or throat clearing sounds.

Researchers can find additional information related to each video (trustworthiness, anxiety ratings, video length, video transcriptions…etc.), as well as information regarding the individuals featured in the video clips (attractiveness, age, race…etc.). As the Miami University Deception Detection Dataset (MU3D) dataset was unlabeled, we tried to label it automatically by making use of the information provided in the codebook. After various experiments with different equations and thresholds we found that the highest accuracies were achieved using a threshold of 70% for a parameter called ‘Truthprop’ (which measures the percentage of people who thought the video was deceptive). The videos that scored a Truth Prop of over 70% were labeled as truthful and the ones that didn’t were labeled as deceptive.

## 3. Proposed Solution:

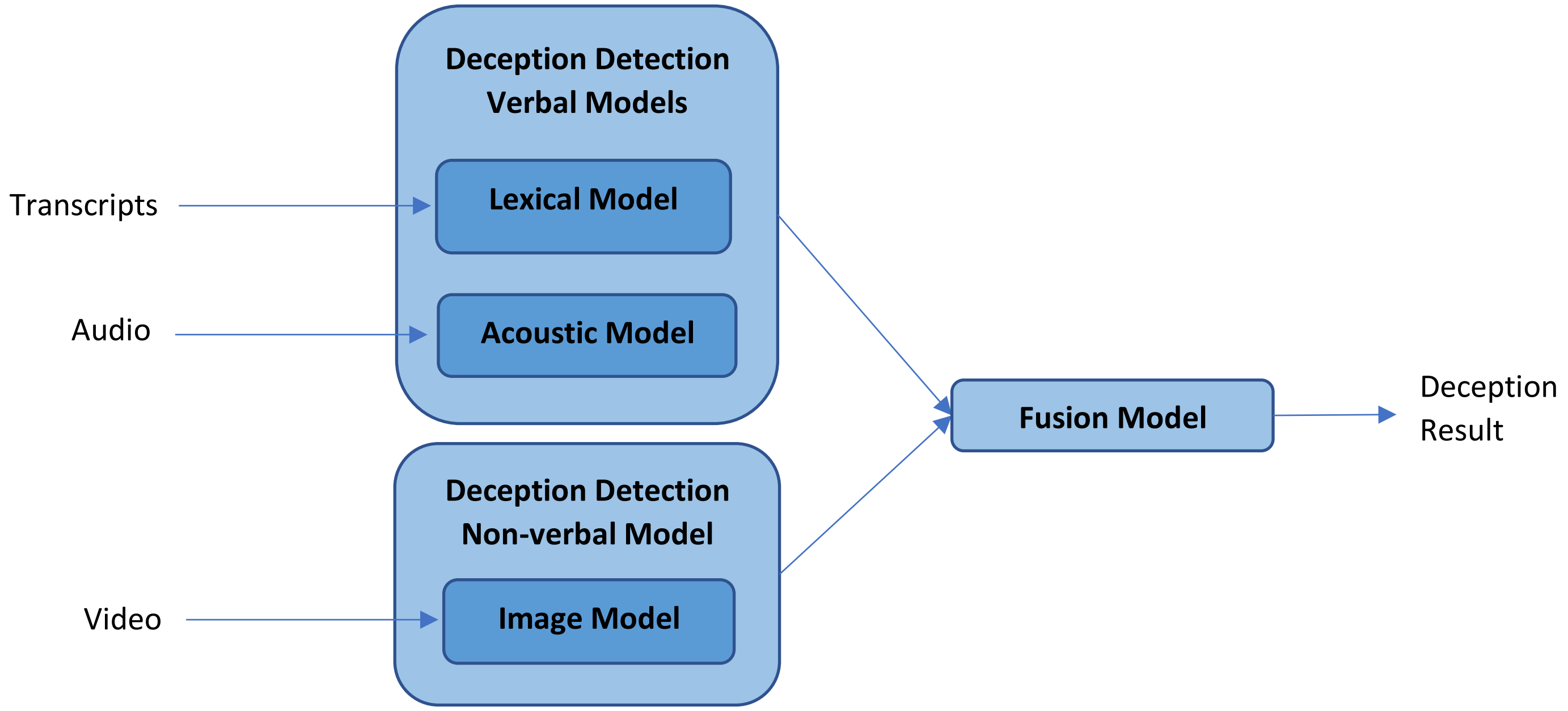

*Figure 1: Our Proposed Solution*

We proposed a system that incorporates three key components: visual features, acoustic features and lexical features. For each component, various machine learning experiments were conducted such as Decision trees [11], Naive Bayes classifiers [12], Support vector machines [13], Gradient Boosting [14], Random forests [15] and Neural networks [16].

### *4.1 Lexical component:*

Several deep learning-based experiments and machine leaning-based experiments were conducted.

#### *4.1.1. Preprocessing:*

First, normalization was done by turning all letter to lowercase. Second, all English stop words were removed. Third, Lemmatization and POS tagging were extracted.

#### *4.1.2. Deep learning Model:*

BERT embedding layer followed by a dropout layer and then a dense layer with a sigmoid activation function and Adam optimizer.

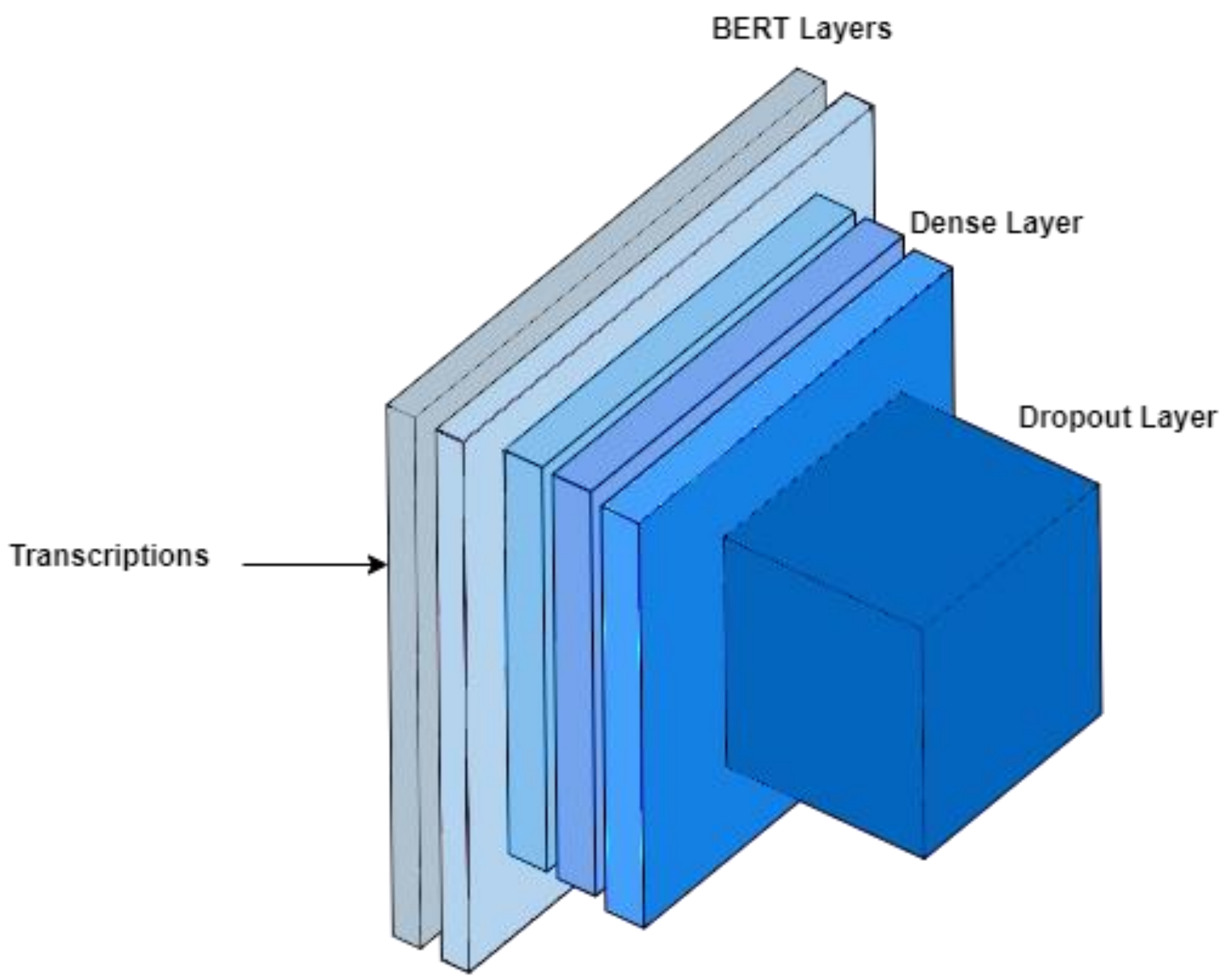

**FIGURE 2: CNN FOR TEXT CLASSIFICATION**

*4.1.3 Support Vector Machine (SVM) Model:*

We proposed using Word2Vec TF-IDF with a Support Vector Machine (SVM) classifier (regularization parameter (C) = 2, coefficient= 9 and degree= 3)

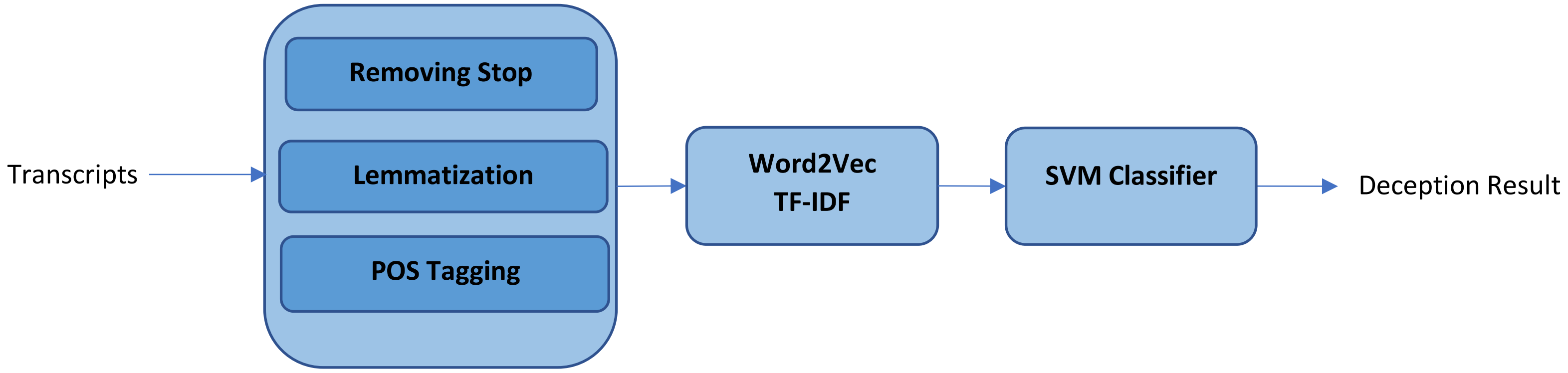

**FIGURE 3: PROPOSED LEXICAL MODEL**

*4.2 Acoustic component:*

*4.2.1. Preprocessing:*

As for the deep learning Model, the audio was clipped into one-second chunks. Then, we standardized then converted the clips to have the exact same sample rate, that way all of the arrays would have equal dimensions. The silence was then padded to increase the duration of the audio and to resized the clips to the same length the next step was data augmentation with time shifting followed by one more round of augmentation but this time instead of being done on the original audio it was done on the Mel spectrogram.

As for the Support Vector Machine (SVM) Model, the clips were resized into four-seconds long frames. Then 25 various features were extracted from the audio clips using the Librosa library [17], including: Chroma STFT, Zero-crossing, RMS, Mel Spectrogram, Roll-off and audio bandwidth

*4.2.2 Deep learning Model:*

A custom data loader was defined and the data was inserted into a model containing 8 convolution layers with Relu activation function, 5 adaptive layers and a linear layer with a learning rate of 0.5.

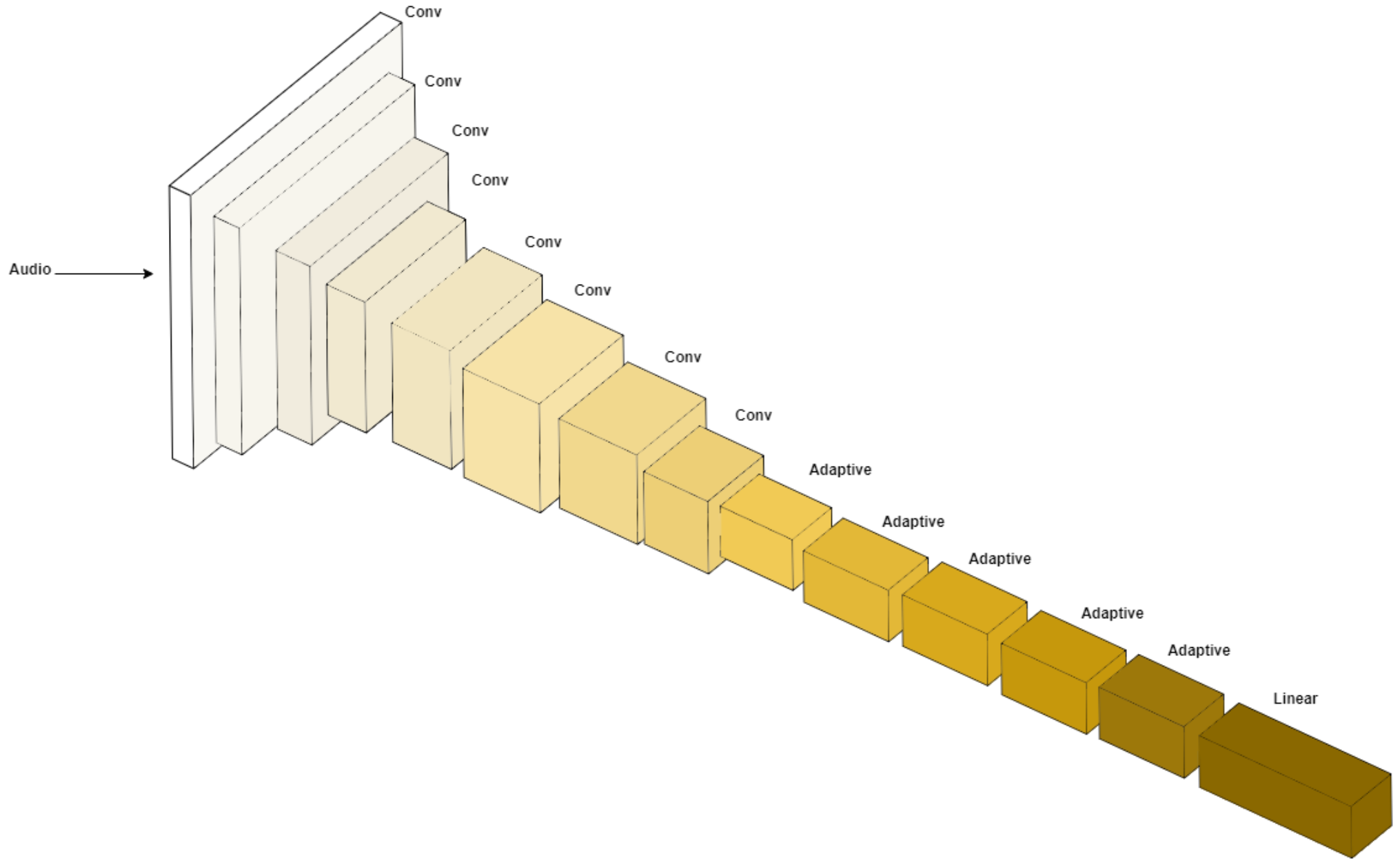

**FIGURE 4: CNN FOR AUDIO CLASSIFICATION**

*4.2.3 Support Vector Machine (SVM) Model:*

After extracting the features, the values of those features were normalized and they were fed to a support vector machine classifier with a regularization parameter (C) of 2 and an RBF type kernel with a coefficient of 6 and a degree of 3.

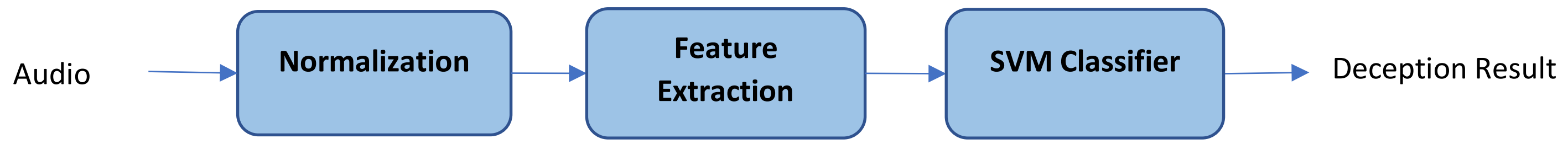

**FIGURE 5: ACOUSTIC MODEL PROPOSED SOLUTION**

*4.3 Visual Component:*

*4.3.1. Preprocessing:*

The proposed solution focuses mainly on the target's facial expressions. Each 0.1 second from each video was turned into a frame in order to get as many samples as possible. The frames were then resized to have the same dimensions.

Face detection was performed using the MTCNN face detection algorithm, and noticed that a lot of the frames contained people that were present during the trial other than the defendant being analyzed (the judge, the security, the audience…) so all of the images which contain more than one face were filtered.

Face detection however was not necessary when dealing with the Mu3d as the quality of the videos was much better and only one individual appeared in each frame, so we were able to obtain good results just by simply using the entire frame.

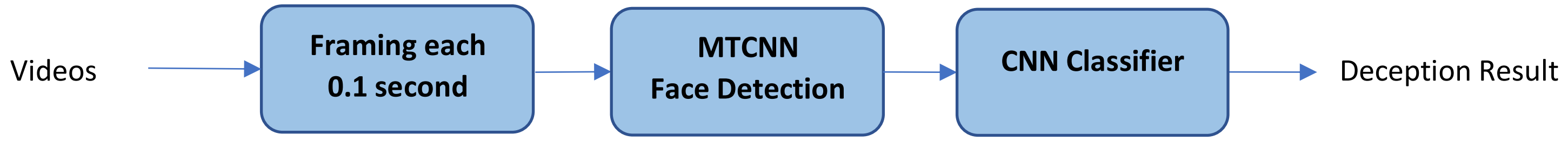

**FIGURE 6 VIDEO MODEL PROPOSED SOLUTION**

*4.3.2. Deep learning Models:*

For our first excitement, we used all of the frames regardless of whether they contain one or several faces. We fed them to a CNN consisting of 4 convolution layers with a Relu activation function followed by a dense layer with a Relu activation function then another dense layer with a sigmoid activation function and Adam optimizer.

For our second excitement, we focused only on the defendant by only detecting faces from frames that contain a single face and feeding them to the same previous model which achieved better results than the previous experiment.

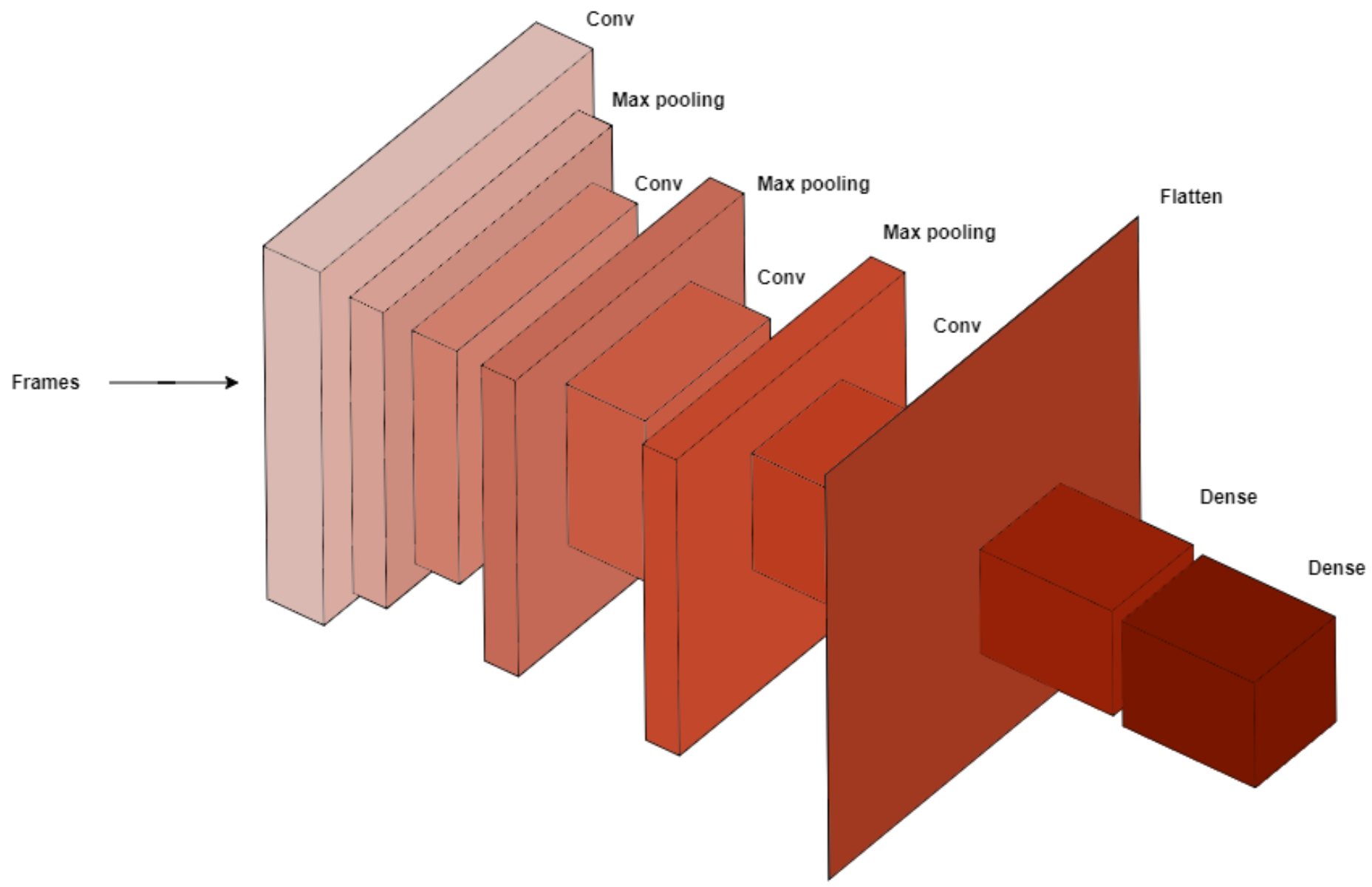

FIGURE 7: CNN FOR VIDEO CLASSIFICATION

## 5. Results and Discussion:

We have compared our results with previous state-of-the-art in the tables 2, 3, 4, 5, then we discussed our experiment results in detail.

### Image Model Results

| | Dataset | Year | Accuracy |
|---|---|---|---|
| **[3]** | Real-life Trial Dataset | 2015 | 75.20% |
| **[4]** | Real-life Trial Dataset | 2019 | 78.95% |
| **[8]** | Real-life Trial Dataset | 2017 | 78.57% |
| **[9]** | Real-life Trial Dataset | 2022 | 61.5% |
| **Our Solution** | Real-life Trial Dataset | 2023 | **97%** |

TABLE 2 IMAGE MODEL RESULTS

### Audio Model Results

| | **Dataset** | **Year** | **Accuracy** |
|---|---|---|---|
| **[8]** | Real-life Trial Dataset | 2017 | 87.5% |
| **[9]** | Real-life Trial Dataset | 2022 | 63.28% |
| **Our Solution** | Real-life Trial Dataset | 2023 | **96%** |

TABLE 3 ACOUSTIC MODEL RESULTS

**Text Model Results**

| | Dataset | Year | Accuracy |
|---|---|---|---|
| **[4]** | Real-life Trial Dataset | 2019 | 66.12% |
| **[8]** | Real-life Trial Dataset | 2017 | 83.78% |
| **[9]** | Real-life Trial Dataset | 2022 | 71.7% |
| **Our Solution** | Real-life Trial Dataset | 2023 | **92%** |

TABLE 4 LEXICAL MODEL RESULTS

**Multi-Modal Model Results**

| | Dataset | Year | Accuracy |
|---|---|---|---|
| **[3]** | Real-life Trial Dataset | 2015 | **-** |
| **[4]** | Real-life Trial Dataset | 2019 | **-** |
| **[8]** | Real-life Trial Dataset | 2017 | 96.42% |
| **[9]** | Real-life Trial Dataset | 2022 | - |
| **Our Solution** | Real-life Trial Dataset | 2023 | **97%** |

TABLE 5 MULTI MODEL RESULTS

*5.1. Lexical component results:*

The best accuracy on text was 92%, achieved using a multinomial naïve Bayes model with default parameters, we also achieve a similar accuracy of 91% using a Support Vector Machine (SVM) model (C=1, Gamma = 9) and notice that lowering gamma or raising C too much results in a reduced accuracy. The best accuracy using the deep learning Model was 75% using the CNN shown in Figure2 on the Real-life Trial dataset and Adam optimizer.

| Support Vector Machine (SVM) | | | |
|---|---|---|---|
| | **C** | **Gamma** | **Accuracy** |
| **Real-Life Trial Dataset** | 1 | 4 | 79% |
| | 1 | 9 | **91%** |
| | 2 | 9 | 82% |
| | 3 | 9 | 74% |
| **Miami University Deception Detection Dataset (MU3D)** | 1 | 3 | 65% |
| | 1 | 9 | **68.75%** |
| | 2 | 9 | 66% |
| | 3 | 9 | 60% |

TABLE 6: LEXICAL MODEL RESULTS USING SUPPORT VECTOR MACHINE (SVM)

On the Miami University Deception Detection Dataset (MU3D) the best accuracy was 73% also using Multinomial Naïve Bayes with default parameters and an accuracy of 68.7% was achieved using a Support Vector Machine (SVM) model (C=1 , Gamma=9).
The deep learning results were less than ideal achieving only 50% using the CNN shown in Figure2 and Adam optimizer.

| Multinomial Naive Bayes Accuracy | |
|---|---|
| **Real-Life Trial Dataset** | **92%** |
| **Miami University Deception Detection Dataset (MU3D)** | **73%** |

**TABLE 7: LEXICAL MODEL RESULTS USING NAIVE BAYES**

*5.2. Acoustic component results:*

Out of all the experiments done on audio, the best results were achieved using Support Vector Machine (SVM) model (C=2, Gamma = 1) which achieved an accuracy of 96% on the Real-life trial dataset. Results using the random forest model showed an accuracy of 84% when max depth set to 4, any depth over 4 resulted in overfitting. Finally, the best accuracy on the Gradient boosting model was 88% (Number of estimators = 50, Learning Rate = 1, Max Depth =1, gamma=4)

The best accuracy using the deep learning Model was 61% using the CNN (Batch size: 32, learning rate 0.01) on the real-life trial dataset.

The best result on Miami University Deception Detection Dataset (MU3D) was accuracy of 82% using the gradient boosting model (number of estimators = 5, learning rate = 0.5, max depth = 1).

The best accuracy using the deep learning Model was 60% with high loss using the CNN shown in Figure 4 on the Miami University Deception Detection Dataset (MU3D).

| Support Vector Machine (SVM) | | | |
|---|---|---|---|
| | **C** | **Gamma** | **Accuracy** |
| **Real-Life Trial Dataset** | 3 | 1 | 96% |
| | 2 | 1 | **96%** |
| | 4 | 1 | 97% |
| **Miami University Deception Detection Dataset (MU3D)** | 3 | 1 | 52% |
| | 2 | 1 | **53%** |
| | 4 | 1 | 52% |

**TABLE 8: ACOUSTIC MODEL RESULTS USING SUPPORT VECTOR MACHINE (SVM)**

| Random Forest | | |
|---|---|---|
| | **Max Depth** | **Accuracy** |
| **Real-Life Trial Dataset** | 2 | 79% |
| | 3 | 82% |
| | 4 | **84%** |
| **Miami University Deception Detection Dataset (MU3D)** | 2 | **57%** |
| | 3 | 56% |
| | 4 | 54% |

**TABLE 9: ACOUSTIC MODEL RESULTS USING RANDOM FOREST**

| Gradient Boosting | | | | |
|---|---|---|---|---|
| | **Num of Estimators** | **Learning Rate** | **Max depth** | **Accuracy** |
| **Real-Life Trial Dataset** | 100 | 1.0 | 1 | 90% |
| | 50 | 1.0 | 1 | **88%** |
| | 10 | 0.5 | 1 | 81% |
| | 10 | 0.1 | 3 | 84% |
| | 20 | 0.3 | 5 | 93% |
| | 5 | 0.1 | 1 | 82% |
| **Miami University Deception Detection Dataset (MU3D)** | 100 | 1.0 | 1 | 53% |
| | 50 | 1.0 | 1 | 53% |
| | 10 | 0.5 | 1 | 81% |
| | 10 | 0.1 | 3 | 52% |
| | 20 | 0.3 | 5 | 52% |
| | 5 | 0.1 | 1 | **82%** |

**TABLE 10: ACOUSTIC MODEL RESULTS USING GRADIENT BOOSTING**

### *5.3 Visual component results:*

The best results were obtained using feature extraction algorithm that filter out any irrelevant faces that didn’t belong to the defendant. Using a CNN with 6-layer convolutional neural network shown above with Adam optimizer.

Results on the Miami University Deception Detection Dataset (MU3D) were 97% using the full pictures. Face detection was not needed when dealing with the Mu3d as the quality of the videos was much better than Miami and only one individual appeared in each frame, so we were able to obtain good results just by simply using the entire frame.

| CNN Accuracy | | |
|---|---|---|
| **Real-Life Trial Dataset** | **With filtering unrelated faces (committee and audiences' faces)** | 95% |
| | **Without filtering unrelated faces (committee and audience faces)** | **97%** |

**TABLE 11: VIDEO MODEL RESULTS WITH AND WITHOUT UNRELATED FACES**

## 6. Conclusion:

We proposed a voting-based method for automatic deception detection using verbal and non-verbal features, machine learning and deep learning. We implemented a voting on results from different lexical, acoustic and visual models on dataset of videos in order to achieve the best accuracies. Our proposed solution outperforms previous state-of-the-art models. Our Voting-based multimodal proposed solution consists of three models. The first model is CNN for detecting deception from images, the second model is Support Vector Machine (SVM) on Mel spectrograms for detecting deception from audio and the third model is Word2Vec on Support Vector Machine (SVM) for detecting deception from manuscripts. Experiments were conducted on Miami dataset and Miami University Deception Detection Dataset (MU3D) dataset. Best results achieved on images, audio and text were 97%, 96%, 92% respectively on Real-Life Trial Dataset, and 97%, 82%, 73% on video, audio and text respectively on Miami University Deception Detection Dataset (MU3D). Using the fusion equation which is (audio model results + image model results + text model results), we achieved an overall accuracy of around 90% on all 3 models using the real-life trial dataset and 77% on the Miami University Deception Detection Dataset (MU3D).

## 7. Declarations:

### Availability of data and materials

All datasets in this survey are available online, you can find links in references.

### Abbreviations

**CNN**: Convolutional Neural Network

**SVM**: Support Vector Machine

**MU3D**: Miami University Deception Detection Dataset

**Acknowledgements**

This paper and the research behind it would not have been possible without the exceptional support of our supervisors. We would like to express out deep gratitude to Professor Khloud Al Jallad for her support and guidance throughout this project. She was without a doubt the reason we were able to finish this work through her constant encouragement and willingness to give her time so generously. Also, Professor Anas Dahabiah, for his patient guidance, enthusiastic encouragement and useful critiques of this research work. We are thankful for their comments on earlier version of the manuscript.

Many thanks to everyone at Arab International University, staff and professors for their incredible support and kind guidance during our time there, we also extend a thanks to all of our classmates for their encouragement and moral support.

**Funding**

The authors declare that they have no funding.

**Author information**

**Authors and Affiliations**

Faculty of Information Technology, Arab International University. Daraa, Syria.

**Contributions**

Lana Touma took on the main role of text models so she performed the literature review, conducted the experiments and wrote the manuscript. Mohammad A- Horani, took on the main role of image models so he performed the literature review, and conducted the experiments as well as helping with the audio experiments. Manar Tailouni took on the main role of audio models so he performed the literature review, and conducted the experiments. Anas Dahabiah and Khloud Al Jallad took on a supervisory role, they made contribution to the conception and analysis of the work and oversaw the completion of the work.

All authors read and approved the final manuscript.

**Ethics declarations**

**Ethics approval and consent to participate**

The authors Ethics approval and consent to participate.

**Consent for publication**

The authors consent for publication.

**Competing interests**

The authors declare that they have no competing interests.